\documentclass[sigconf]{acmart}

\usepackage{amsmath}
\usepackage{multirow}
\usepackage{booktabs}
\usepackage{graphicx}
\usepackage{enumitem}

\AtBeginDocument{%
  }

\setcopyright{acmlicensed}
\copyrightyear{2026}
\acmYear{2026}
\acmDOI{XXXXXXX.XXXXXXX}
\acmConference[MILETS '26]{The 12th International Workshop on Mining and
  Learning from Time Series}{August 10, 2026}{Jeju, Republic of Korea}

\acmBooktitle{Proceedings of the 12th International Workshop on Mining
and Learning from Time Series (MILETS '26), August 10, 2026, Jeju,
Republic of Korea}

\acmISBN{978-1-4503-XXXX-X/2026/08}

\begin{document}

\title{Missingness as Signal: Channel-Independent Spectrogram Learning for Clinical Time Series Prediction}

\author{Soyeon Park}
\email{soyeon.park@handong.ac.kr}
\orcid{0009-0008-9560-8583}
\affiliation{%
  \institution{Handong Global University}
  \city{Pohang}
  \country{Republic of Korea}
}
\author{Charmgil Hong}
\email{charmgil@handong.ac.kr}
\orcid{0000-0002-8176-252X}
\affiliation{%
  \institution{Handong Global University}
  \city{Pohang}
  \country{Republic of Korea}
}


\begin{abstract}
Clinical time series prediction in intensive care units remains challenging due to heterogeneous physiological variables and informative missingness.
The presence or absence of a measurement can reflect clinical decisions and patient severity, and thus missingness can serve as a predictive signal rather than a simple data artifact.
This work presents CISM, a Channel-Independent Spectrogram framework with a Missingness stream for clinical multivariate time series prediction.
CISM converts each clinical variable into a variable-wise time-frequency spectrogram, preserves variable identity through variable-aligned encoding, and aligns an explicit missingness stream with the spectrogram representation.
Experiments on an in-hospital mortality task derived from MIMIC-IV show that CISM achieves the highest mean AUROC (0.7225), AUPRC (0.3308), and F1 (0.3808) among the compared time series, missingness-aware, vision, and time-frequency baselines.
Ablation studies further show that observation patterns provide a meaningful informative signal.
Pixel-level mask injection improves performance over plain spectrogram inputs and recovers much of this predictive value. 
The aligned missingness stream contributes a further, complementary gain in both AUROC and AUPRC.
These results highlight the importance of modeling observation patterns as structured signals in clinical time series prediction.
\end{abstract}




\maketitle

\section{Introduction}
Clinical time series prediction has become an important task in healthcare analytics~\cite{10.1145/3531326}.
Early risk estimation in intensive care units (ICU) can support timely clinical intervention, resource allocation, and patient management~\cite{knaus1985apache, che2018recurrent}.
In-hospital mortality prediction is a representative setting for this problem. The task aims to estimate the mortality risk of a patient from physiological measurements collected during the early stage of ICU admission.

Clinical multivariate time series differ from regular sensor data in two demanding respects~\cite{che2018recurrent, harutyunyan2019multitask}.
The first is irregular sampling.
ICU measurements are highly irregular across variables: vital signs such as heart rate, blood pressure, oxygen saturation, and respiratory rate are recorded frequently~\cite{PhysioNet-mimiciv-2.0}, whereas laboratory-related variables such as pH, glucose, temperature, and fraction of inspired oxygen appear much less often.
In our cohort, the missing rate ranges from approximately 3.6\% for heart rate to 89.7\% for pH, which indicates that missingness is not a uniform data artifact.
The second is variable heterogeneity.
ICU variables differ in scale, unit, physiological meaning, and temporal behavior, and early fusion across all variables may obscure these differences.
This issue becomes more severe under variable-specific missingness, as each variable follows its own measurement pattern.

These two properties make ICU time series difficult to model.
However, missingness is not only a nuisance.
The presence or absence of a measurement often reflects a clinical decision, patient severity, or treatment context~\cite{groenwold2020informative, tan2023informative}.
For example, a recorded pH value may indicate that arterial blood gas analysis was performed, and a recorded FiO\(_2\) value may indicate respiratory support.
The observation pattern itself can provide predictive information, and a model that treats missing values only as noise to impute may discard part of the clinical signal~\cite{lipton2016modeling}.
This view motivates a model that exploits irregularity and heterogeneity as structure, and that uses observation patterns as informative input rather than discarding them through imputation.

Existing approaches address these challenges only in part.
Time series models such as recurrent networks, temporal convolutional networks, and Transformers usually operate directly on raw temporal sequences~\cite{zhou2021informer, wu2022timesnet}.
Some models include an observation mask as an auxiliary input channel; this exposes missingness information, though the mask itself remains a one-dimensional temporal feature aligned with the raw sequence~\cite{che2018recurrent}.
Spectrogram-based approaches instead convert temporal signals into time-frequency images~\cite{huzaifah2017comparison}. Standard vision backbones, however, process the resulting spectrogram as a single image, which can ignore variable boundaries and mix heterogeneous clinical variables at early layers.

This work proposes CISM, a Channel-Independent Spectrogram framework with a Missingness stream for clinical multivariate time series prediction.
To represent informative missingness, CISM uses a parallel missingness stream in which the observation mask is rendered as a one-channel image with the same spatial geometry as the spectrogram.
Each mask patch corresponds to a spectrogram patch, which enables patch-level alignment between the two streams.
The two streams are fused at the patch level, and the model combines physiological time-frequency patterns with local observation patterns.
This design treats missingness as an informative predictive signal rather than a preprocessing artifact.

The proposed model is evaluated on a MIMIC-derived in-hospital mortality task with ten clinical variables from the first 48 hours of ICU admission.
It improves on dedicated missingness-aware baselines for irregular and incomplete clinical time series, and it significantly outperforms the time series, vision, and time-frequency baselines.
Ablation studies indicate that observation patterns provide a meaningful predictive signal: pixel-level masking improves performance over plain spectrogram inputs and yields the larger share of the gain, and the aligned missingness stream adds a further, complementary improvement in both AUROC and AUPRC.

The contributions of this work are summarized as follows.
\begin{itemize}[leftmargin=*]
    \item We propose CISM, a Channel-Independent Spectrogram framework with a Missingness stream that preserves variable identity through variable-aligned encoding and exploits time-frequency representations derived from individual variables.
    \item We introduce an explicit missingness stream aligned with the spectrogram representation, which allows the model to treat observation patterns as predictive clinical signals. 
    \item We provide a unified comparison against time series, missingness-aware, vision, and time-frequency baselines.
    \item We conduct ablation studies that isolate the effects of pixel-level missingness injection, the explicit missingness stream, spectrogram-missingness fusion, and wavelet transform choice.
\end{itemize}

\section{Related Work}

\subsection{Clinical Time Series Prediction}
Clinical time series prediction has been widely studied for ICU risk estimation, disease progression analysis, and patient outcome prediction~\cite{harutyunyan2019multitask}.
In-hospital mortality prediction is a representative task in this area. 
The objective is to estimate patient risk from physiological measurements collected during the early stage of ICU admission.
Prior studies have explored recurrent neural networks, temporal convolutional networks, and Transformer-based architectures that operate on raw temporal sequences and learn temporal dependencies directly from observed values~\cite{harutyunyan2019multitask, lipton2016modeling}.

Recent general-purpose time series models, such as PatchTST~\cite{nie2022time}, Informer~\cite{zhou2021informer}, FEDformer~\cite{zhou2022fedformer}, Pyraformer~\cite{liu2021pyraformer}, and TimesNet~\cite{wu2022timesnet}, provide strong encoders for temporal data and perform competitively in forecasting and classification settings.
However, clinical ICU data differ substantially from regular multivariate sensor data: measurements are irregular across variables, missingness patterns vary substantially, and physiological variables have distinct units, scales, and clinical meanings.
Raw sequence models may have limited ability to exploit variable-specific structure and observation patterns in ICU data without explicit variable separation~\cite{lipton2016modeling}.

\subsection{Time Series as Spectrograms and Time-Frequency Models}
Another line of research transforms time series into time-frequency representations.
Spectrograms provide two-dimensional views of temporal signals, in which local temporal variation and frequency structure are represented jointly~\cite{daubechies1990wavelet, sejdic2009time}.
Such representations, widely used in audio analysis and explored for general time series prediction, offer a natural way to apply image encoders to temporal data~\cite{zeng2023pixels}.
The Audio Spectrogram Transformer (AST)~\cite{gong2021ast} applies Transformer architectures to spectrogram inputs, whereas MultiWave~\cite{deznabi2023multiwave} and SpecAR-Net~\cite{ijcai2024p433} use wavelet or spectral representations for temporal modeling.
These approaches motivate the use of time-frequency representations for time series prediction with vision-based neural architectures.

However, direct application of spectrogram-based models to clinical multivariate time series has an important limitation~\cite{ghassemi2015multivariate}.
A stacked spectrogram contains multiple clinical variables, each with a distinct physiological meaning and missingness profile.
Standard vision backbones process the entire spectrogram as a single image, without an explicit constraint that preserves variable boundaries.
As a result, heterogeneous variables can be mixed at early layers.
Existing time-frequency models also do not explicitly align missingness patterns with variable-wise spectrogram regions.
Our work addresses these limitations through variable-aligned spectrogram encoding and an explicit missingness stream.

\subsection{Informative Missingness and Channel-Independent Modeling}
Missingness is a central property of clinical time series~\cite{che2018recurrent, groenwold2020informative}.
In ICU data, the presence or absence of a measurement can reflect clinical decisions, disease severity, and treatment context~\cite{groenwold2020informative, PhysioNet-mimiciv-2.0}.
Prior methods address missingness through imputation, temporal decay, observation masks, and irregular sampling mechanisms.
GRU-D employs trainable decay terms and observation masks for multivariate clinical time series with missing values~\cite{che2018recurrent}, while mTAND and SeFT exploit observation patterns or set-based representations for sparse and irregular measurements~\cite{shukla2021multi, horn2020set}.
Collectively, these approaches establish that missingness conveys predictive information rather than pure noise~\cite{lipton2016modeling}.
However, they represent missingness as a one-dimensional temporal signal over the raw sequence or as an irregular observation pattern in the original time domain, and therefore do not model missingness in a time-frequency space.

Channel-independent modeling provides another relevant direction for multivariate time series~\cite{nie2022time, han2024capacity}.
Rather than mix all variables at the input level, channel-independent models process each variable separately and aggregate variable-level representations at a later stage, which reduces harmful early interactions among heterogeneous variables.
Prior work has mainly applied this idea to raw temporal sequences.
In contrast, our work extends a variable-aligned form of channel independence to time-frequency representations, encoding each variable band separately and aligning missingness through a dedicated spatial stream.

\begin{figure*}[t]
    \centering
    \includegraphics[width=\textwidth]{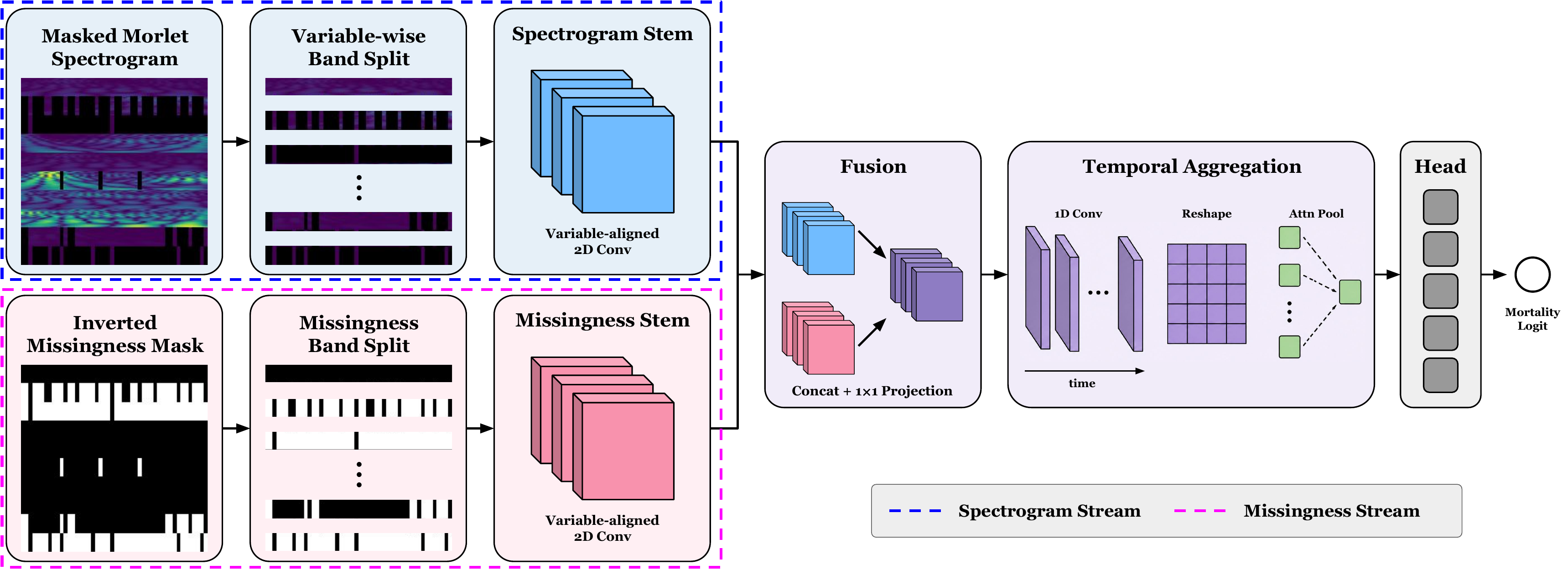}
    \caption{Overall architecture of CISM.
    Each clinical variable is converted into a time-frequency spectrogram band, and the observation mask is rendered as an aligned missingness image.
    The two streams are encoded with variable-aligned stems, fused at the patch level, and aggregated for mortality prediction.}
    \label{fig:framework}
\end{figure*}

\section{Channel-Independent Spectrogram Framework with a Missingness Stream}
We propose CISM, a Channel-Independent Spectrogram framework with a Missingness stream for clinical multivariate time series prediction.
The framework converts each clinical variable into a time-frequency spectrogram and preserves variable identity through variable-aligned encoding.
Missingness is represented through two pathways: pixel-level masking in the spectrogram image and a dedicated missingness stream aligned with the spectrogram.
The model fuses spectrogram and mask features at the patch level, then aggregates variable-specific representations for mortality prediction.
Figure~\ref{fig:framework} illustrates the overall architecture of CISM.
The main design principle is to delay cross-variable interaction until variable-level representations have been obtained.

\subsection{Problem Formulation}
Let \(X_i \in \mathbb{R}^{V \times T}\) denote a multivariate clinical time series for the \(i\)-th ICU stay, where \(V\) is the number of clinical variables and \(T\) is the length of the observation window, and let \(x_{i,v,t}\) be the value of variable \(v\) at time step \(t\).
Each series is associated with an observation mask \(M_i \in \{0,1\}^{V \times T}\), where \(m_{i,v,t}=1\) indicates an observed value and \(m_{i,v,t}=0\) a missing value.
The missingness stream uses the inverted mask \(\bar{M}_i = 1 - M_i\), where \(\bar{m}_{i,v,t}=1\) denotes a missing entry, and the same observation pattern suppresses the corresponding regions in the masked spectrogram input.
Each ICU stay has a binary label \(y_i \in \{0,1\}\), where \(y_i=1\) denotes in-hospital mortality.
The objective is to learn a prediction function \(f\) that maps the clinical time series and its observation pattern to a mortality probability \(\hat{y}_i = f(X_i, M_i)\).
The prediction target is defined at the ICU-stay level rather than at the variable level, and thus the model learns patient-level risk from both physiological values and observation patterns under a binary classification objective.

\begin{figure}[t]
    \centering
    \includegraphics[width=\columnwidth]{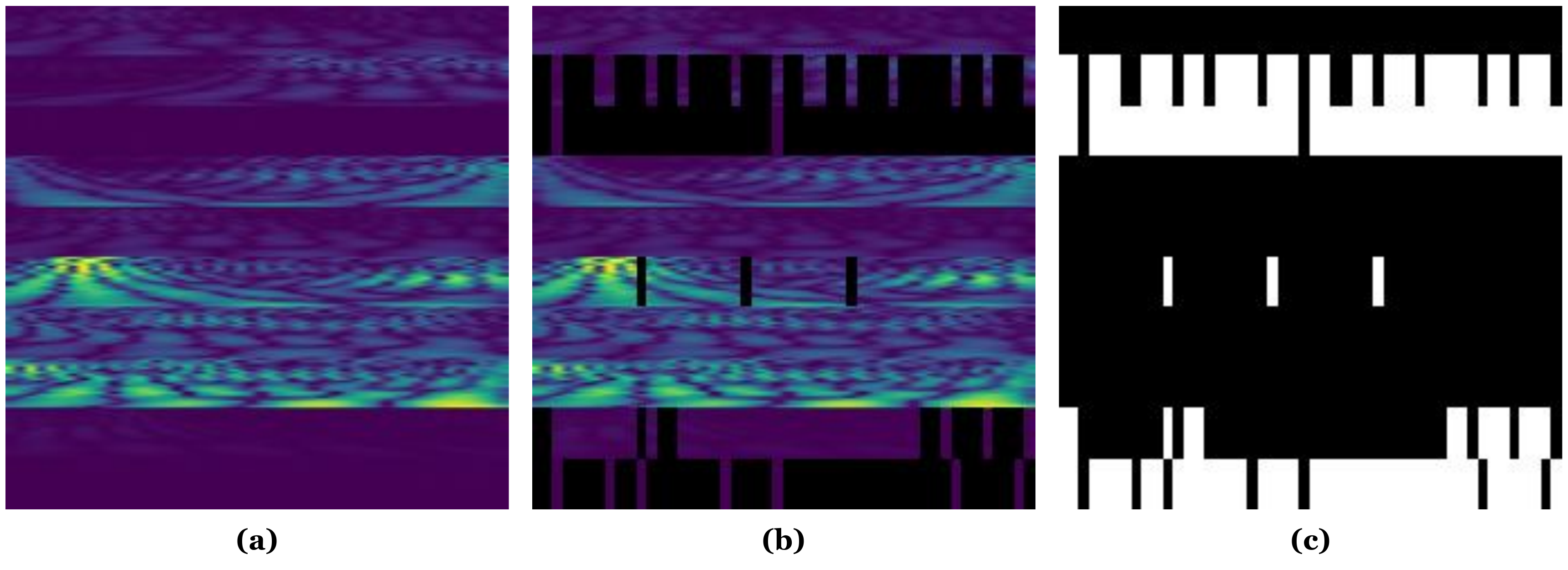}
    \caption{Example input representations used by CISM.
    (a) Unmasked variable-wise spectrogram.
    (b) Masked spectrogram with missing regions suppressed at the pixel level.
    (c) Aligned missingness image from the inverted observation mask, where missing regions are shown in white.}
    \label{fig:input_representation}
\end{figure}

\subsection{Variable-Wise Spectrogram Construction}
Each clinical variable is independently transformed into a time-frequency representation through a continuous wavelet transform.
Missing entries are filled under the dataset protocol and the transform is applied to the imputed time series.
The Morlet wavelet serves as the default transform, owing to its localized time-frequency representation~\cite{daubechies1990wavelet} and its stable empirical performance in the ablation study.
The Complex Morlet wavelet is evaluated as an alternative to examine the effect of wavelet choice.

The time-frequency map of each variable is rendered as an RGB spectrogram band with a fixed height and temporal width.
This choice ensures compatibility with image-based baselines and pretrained vision backbones, which typically require three-channel inputs.
This design provides a consistent input format across CISM, vision backbones, and time-frequency baselines.
A direct comparison between RGB-rendered spectrograms and raw single-channel CWT coefficients remains a direction for future study.

Pixel-level masking is applied with the original observation mask, which is mapped to the spatial layout of the spectrogram such that regions associated with missing entries are suppressed in the corresponding variable band.
Since wavelet coefficients reflect neighboring temporal values, this post-hoc masking may not fully remove the influence of imputed values on adjacent coefficients.
Pixel-level masking therefore serves as a conservative constraint on the spectrogram input, whereas the original observation mask is retained for the missingness stream described in the next subsection.
Finally, the variable-wise bands are stacked vertically into a single spectrogram image, in which each variable occupies a fixed row region that keeps variable boundaries explicit and preserves variable identity in an image representation.

\subsection{Missingness Stream Input}
The missingness stream uses a separate mask image derived from the original observation mask.
The mask $M_i$ is first inverted to $\bar{M}_i$, where missing entries are set to one and observed entries to zero, and the inverted mask is rendered with the same spatial layout as the spectrogram image.
Each mask region is therefore aligned with the corresponding variable-wise spectrogram region, which allows the model to associate local time-frequency patterns with local missingness patterns.
This aligned mask image serves as the input to the missingness stream.

\subsection{Channel-Independent Two-Stream Encoder}

The encoder consists of a spectrogram stream and a missingness stream.
The spectrogram stream receives the masked spectrogram image as its main input.
Its convolutional stem uses kernels whose height matches the height of one variable band.
The vertical stride is set to the same height, and each kernel covers one variable band at a time.
This design prevents early convolutional operations from crossing variable boundaries.
As a result, variable identity is preserved during early representation learning.

The missingness stream receives the aligned mask image and follows the same kernel geometry as the spectrogram stream.
This shared geometry creates a one-to-one correspondence between spectrogram patches and mask patches.
Each paired patch represents the same clinical variable and the same temporal region.
As a result, the two streams produce aligned feature sequences before fusion.
This structure allows missingness information to be modeled as an explicit signal rather than as an auxiliary scalar feature.
It keeps the missingness representation separate from the physiological representation until patch-level fusion.

\subsection{Patch-Level Fusion}
After the two stems produce aligned patch features, the corresponding spectrogram and mask features are fused at each patch.
The main configuration uses concatenation followed by a learnable projection.
This fusion combines local time-frequency information with the corresponding local missingness pattern.
The resulting fused features form a variable-specific temporal sequence.
Alternative fusion strategies, such as addition, gating, and FiLM-based modulation, are examined in the ablation study.
The main model uses concatenation due to its stable empirical performance.

\subsection{Variable-Level Aggregation and Prediction}
The fused patch sequence of each variable is processed by a shared temporal trunk, and temporal pooling then summarizes it into a variable-level representation derived from both spectrogram and missingness features.
Cross-variable interaction is introduced only after these variable-level representations have been obtained.
This delayed aggregation reduces premature mixing among heterogeneous clinical variables and retains variable-specific information before patient-level prediction.
The variable-level representations are then aggregated through attention pooling, which assigns variable-level importance and produces a patient-level representation.
A prediction head maps this representation to a mortality probability. The model is trained with a binary classification objective.

\section{Experiments}

\begin{table}[t]
\centering
\caption{Variable-wise missing rates (\%) across train, validation, and test splits.}
\label{tab:missing_rates}
\begin{tabular*}{\columnwidth}{@{\extracolsep{\fill}}lcccc}
\toprule
Variable & Train & Val & Test & Overall \\
\midrule
Diastolic BP & 5.7 & 5.6 & 5.6 & 5.7 \\
FiO\(_2\) & 83.3 & 83.6 & 83.2 & 83.3 \\
Glucose & 77.3 & 76.8 & 76.7 & 77.1 \\
Heart Rate & 3.6 & 3.4 & 3.5 & 3.6 \\
Mean BP & 5.7 & 5.5 & 5.5 & 5.6 \\
Oxygen Saturation & 5.0 & 4.4 & 4.7 & 4.9 \\
Respiratory Rate & 3.9 & 3.7 & 3.8 & 3.9 \\
Systolic BP & 5.7 & 5.6 & 5.5 & 5.7 \\
Temperature & 70.9 & 69.7 & 70.7 & 70.7 \\
pH & 89.8 & 89.2 & 89.2 & 89.7 \\
\bottomrule
\end{tabular*}
\end{table}

\begin{table*}[t]
\centering
\caption{Main comparison across model families on in-hospital mortality prediction.
Time series baselines use ten clinical variables with inverted missing indicator channels.
Missingness-aware baselines explicitly model missing values or irregular observations.
Vision and time-frequency baselines use Morlet spectrogram images.
Results are reported as mean $\pm$ standard deviation over five seeds.
\textbf{Bold} denotes the best mean performance for each metric and statistically comparable results under the paired bootstrap test
($B = 10{,}000$, $p \geq 0.05$).}
\label{tab:main_results}
\begin{tabular*}{\textwidth}{@{\extracolsep{\fill}}llccc}
\toprule
Family & Model & AUROC & AUPRC & F1 \\
\midrule
\multirow{5}{*}{Time Series}
& FEDformer~\cite{zhou2022fedformer} & 0.6366 $\pm$ 0.0059 & 0.2184 $\pm$ 0.0091 & 0.2475 $\pm$ 0.0071 \\
& Informer~\cite{zhou2021informer} & 0.6778 $\pm$ 0.0027 & 0.2569 $\pm$ 0.0057 & 0.3192 $\pm$ 0.0103 \\
& PatchTST~\cite{nie2022time} & \textbf{0.7001 $\pm$ 0.0064} & 0.2691 $\pm$ 0.0115 & \textbf{0.3490 $\pm$ 0.0222} \\
& Pyraformer~\cite{liu2021pyraformer} & 0.6669 $\pm$ 0.0222 & 0.2522 $\pm$ 0.0145 & 0.3171 $\pm$ 0.0220 \\
& TimesNet~\cite{wu2022timesnet} & 0.6303 $\pm$ 0.0050 & 0.2249 $\pm$ 0.0041 & 0.2611 $\pm$ 0.0149 \\

\midrule
\multirow{3}{*}{Missingness-aware}
& GRU-D~\cite{che2018recurrent} & \textbf{0.7012 $\pm$ 0.0031} & \textbf{0.3223 $\pm$ 0.0221} & \textbf{0.3646 $\pm$ 0.0207} \\
& mTAND~\cite{shukla2021multi} & \textbf{0.7156 $\pm$ 0.0061} & \textbf{0.3201 $\pm$ 0.0046} & \textbf{0.3587 $\pm$ 0.0159} \\
& SeFT~\cite{horn2020set} & \textbf{0.7026 $\pm$ 0.0107} & 0.2856 $\pm$ 0.0098 & \textbf{0.3754 $\pm$ 0.0079} \\
\midrule
\multirow{5}{*}{Vision}
& ConvNeXt-Tiny~\cite{liu2022convnet} & 0.6568 $\pm$ 0.0214 & 0.2620 $\pm$ 0.0269 & 0.3255 $\pm$ 0.0130 \\
& EfficientNet-B0~\cite{tan2019efficientnet} & 0.6483 $\pm$ 0.0190 & 0.2475 $\pm$ 0.0193 & 0.1961 $\pm$ 0.0291 \\
& ResNet-50~\cite{he2016deep} & 0.6180 $\pm$ 0.0142 & 0.2114 $\pm$ 0.0157 & 0.2091 $\pm$ 0.0400 \\
& Swin-Tiny~\cite{liu2021swin} & 0.6698 $\pm$ 0.0179 & 0.2645 $\pm$ 0.0127 & 0.2853 $\pm$ 0.0225 \\
& ViT-Small~\cite{dosovitskiy2020image} & 0.6326 $\pm$ 0.0229 & 0.2393 $\pm$ 0.0214 & 0.2637 $\pm$ 0.0382 \\
\midrule
\multirow{3}{*}{Time-Frequency}
& AST~\cite{gong2021ast} & 0.6917 $\pm$ 0.0073 & 0.2729 $\pm$ 0.0130 & 0.3216 $\pm$ 0.0371 \\
& MultiWave~\cite{deznabi2023multiwave} & 0.6172 $\pm$ 0.0268 & 0.2474 $\pm$ 0.0268 & 0.2545 $\pm$ 0.0495 \\
& SpecAR-Net~\cite{ijcai2024p433} & 0.5599 $\pm$ 0.0131 & 0.1732 $\pm$ 0.0065 & 0.2525 $\pm$ 0.0408 \\
\midrule
Ours
& CISM & \textbf{0.7225 $\pm$ 0.0103} & \textbf{0.3308 $\pm$ 0.0178} & \textbf{0.3808 $\pm$ 0.0116} \\
\bottomrule
\end{tabular*}
\end{table*}

\subsection{Dataset and Evaluation Protocol}

The experiments are conducted on the in-hospital mortality task from the MedMod~\cite{elsharief2025medmod} benchmark, which is derived from MIMIC-IV~\cite{PhysioNet-mimiciv-2.0}.
The cohort construction and preprocessing follow the MIMIC-IV processing pipeline provided by MedMod.
The task is to predict in-hospital mortality from clinical measurements collected within the first 48 hours of ICU admission.
Missing entries are filled with predefined values according to the MedMod preprocessing procedure, while the observation mask is retained to indicate which entries were originally missing.
The dataset contains 6,215 ICU stays, with 4,485 stays for training, 488 stays for validation, and 1,242 stays for testing.
The positive mortality rate is approximately 14--15\%, which indicates a class-imbalanced prediction task.

The absolute performance under this protocol is not directly comparable to MIMIC in-hospital mortality results from studies with larger feature sets or different benchmark definitions.
The MedMod protocol used in this work restricts the input to 10 continuous clinical variables from the first 48 hours of ICU admission.
It does not include categorical clinical variables, diagnosis codes, medication records, procedure information, clinical notes, or a broader set of laboratory measurements.
This restricted input protocol can lead to lower absolute performance than full-feature MIMIC mortality prediction protocols.
The results should be interpreted as performance under a constrained MedMod protocol for sparse multivariate clinical time series, rather than as a direct comparison to full-feature MIMIC mortality prediction studies.

The input contains 10 continuous clinical variables: diastolic blood pressure, fraction of inspired oxygen, glucose, heart rate, mean blood pressure, oxygen saturation, respiratory rate, systolic blood pressure, temperature, and pH.
Table~\ref{tab:missing_rates} reports the variable-wise missing rates across the train, validation, and test splits.
The missingness profile varies substantially across variables, ranging from 3.6\% for heart rate to 89.7\% for pH.
Variables such as heart rate, respiratory rate, oxygen saturation, and blood pressure show relatively low missing rates.
In contrast, pH, fraction of inspired oxygen, glucose, and temperature show substantially higher missing rates.
This protocol provides a natural testbed for models that explicitly use informative missingness.

All models are trained and evaluated with five random seeds.
Results are reported as mean $\pm$ standard deviation across seeds.
The primary evaluation metrics are AUROC, AUPRC, and F1 score.
AUROC and AUPRC evaluate threshold-free discrimination performance.
F1 score is computed with a threshold selected on the validation set, rather than the default threshold of 0.5.
The selected threshold is applied to the test set for final reporting.

\subsection{Baselines}
Baselines are grouped into four families that separate the effects of raw temporal modeling, explicit missingness-aware modeling, generic image modeling, and specialized time-frequency modeling.

Time series baselines receive the 10 clinical variables with inverted missing indicator channels, which exposes raw sequence models to explicit missingness information and serves as a direct comparison to the proposed missingness stream.
This group comprises FEDformer~\cite{zhou2022fedformer}, Informer~\cite{zhou2021informer}, PatchTST~\cite{nie2022time}, Pyraformer~\cite{liu2021pyraformer}, and TimesNet~\cite{wu2022timesnet}.
Missingness-aware baselines target irregular or incomplete clinical time series.
GRU-D~\cite{che2018recurrent} combines trainable temporal decay terms with observation masks, mTAND~\cite{shukla2021multi} models irregular sampling through time-aware attention, and SeFT~\cite{horn2020set} represents observations as permutation-invariant sets.
These baselines provide direct comparisons with models that explicitly target missingness or irregular sampling.

Vision baselines apply standard image backbones to the same masked Morlet spectrogram images used by CISM, which tests whether generic image encoders alone suffice on stacked clinical spectrograms.
This group comprises ConvNeXt-Tiny~\cite{liu2022convnet}, EfficientNet-B0~\cite{tan2019efficientnet}, ResNet-50~\cite{he2016deep}, Swin-Tiny~\cite{liu2021swin}, and ViT-Small~\cite{dosovitskiy2020image}, all of which process the entire stacked spectrogram as a single image without variable-aligned encoding or a separate missingness stream.
Time-frequency baselines test whether specialized spectral architectures capture the relevant structure in sparse clinical multivariate time series.
This group comprises AST~\cite{gong2021ast}, MultiWave~\cite{deznabi2023multiwave}, and SpecAR-Net~\cite{ijcai2024p433}.
All baselines are evaluated under exactly the same data splits, metrics, and five-seed evaluation protocol as CISM.

\subsection{Implementation Details}
Each continuous variable is standardized per channel through z-score normalization with training-split statistics and transformed into a Morlet spectrogram through a continuous wavelet transform at one-hour temporal resolution.
Each variable yields a fixed-height band, and the ten bands are stacked vertically into a single image; the inverted observation mask is resized to the same layout and forms a one-channel stream aligned with the spectrogram.
All models use AdamW (learning rate $10^{-4}$, weight decay $10^{-2}$) with a cosine schedule, batch size 32, and 30 epochs.
The objective is a class-weighted binary cross-entropy loss, in which the positive class is weighted by the training negative-to-positive ratio, with gradient clipping at 1.0.

We select the checkpoint with the best validation AUROC and report the mean and standard deviation over five seeds.
To test whether performance gaps are statistically significant, we further apply a \emph{paired bootstrap}~\cite{efron1994introduction} to the per-patient test predictions ($B = 10{,}000$ resamples).
Each resample evaluates all models on the same patients and seeds.
A method is marked as not significantly different from the best-performing method when the paired bootstrap yields p $\geq$ 0.05 for the metric difference.
CISM remains lightweight despite the two-stream design, with 0.364M parameters and a throughput of approximately 69K samples per second.

\subsection{Main Results}
Table~\ref{tab:main_results} presents the main comparison across model families.
CISM attains the highest mean AUROC (0.7225), AUPRC (0.3308), and F1 (0.3808) among all compared models.
Among the time series baselines, PatchTST is the strongest, with an AUROC of 0.7001 and an F1 score of 0.3490.
These baselines receive inverted missing indicator channels, although raw sequence models benefit only partially from this explicit missingness information. 
This comparison suggests that one-dimensional auxiliary channels alone are insufficient.

Among the missingness-aware baselines, mTAND attains the highest AUROC (0.7156), GRU-D the highest AUPRC (0.3223), and SeFT the highest F1 (0.3754).
CISM achieves the best mean across all three metrics and is competitive with or better than these missingness-aware baselines.
Since the task is class-imbalanced, AUPRC and F1 are the more informative criteria, and CISM attains the highest mean on both.
The AUROC margin over mTAND is small relative to the seed-level standard deviation, and the paired-bootstrap analysis shows that CISM is not significantly different from the strongest missingness-aware baselines, including mTAND and GRU-D. 
We read these results as evidence that CISM matches or improves on dedicated missingness-aware models, rather than as a decisive gap. 
In contrast, CISM significantly outperforms the time series, vision, and time-frequency baselines under the reported criteria.

The vision baselines test whether the gain stems from spectrogram images alone.
Swin-Tiny is the strongest in this group (AUROC 0.6698), although it remains below CISM on all primary metrics.
A likely reason is that generic image backbones process the spectrogram as a single image without explicit variable boundaries, whereas CISM preserves variable identity through variable-aligned encoding before patient-level aggregation.

CISM also outperforms the time-frequency baselines, among which AST is the strongest (AUROC 0.6917, AUPRC 0.2729).
This result indicates that the gain stems from combining the time-frequency representation with variable-aligned encoding and a dedicated missingness stream, rather than from the representation alone.
We position CISM as a new representation perspective on clinical missingness rather than a state-of-the-art contender on this benchmark: it casts informative missingness as a spatially aligned stream over variable-wise time-frequency bands, rather than a one-dimensional auxiliary channel.
The value of this view lies in how missingness and physiological structure are represented, not in a large margin in absolute terms on a single imbalanced task.
A detailed component-level analysis follows in the ablation study.

\subsection{Ablation Studies}
The ablation study examines the main design choices of CISM: pixel-level missingness injection, the explicit missingness stream, spectrogram-missingness fusion, and wavelet transform choice.
Here, ``unmasked'' denotes a spectrogram without pixel-level missingness suppression, whereas ``masked'' denotes one in which regions associated with missing values are suppressed.
All ablation results are reported as mean \(\pm\) standard deviation over five seeds.

\begin{table}[t]
\centering
\caption{Effect of pixel-level missingness injection without the missingness stream.}
\label{tab:ablation_pixel_mask}
\begin{tabular*}{\columnwidth}{@{\extracolsep{\fill}}lcc}
\toprule
Setting & AUROC & AUPRC \\
\midrule
Morlet, masked & 0.7096 $\pm$ 0.0111 & 0.3218 $\pm$ 0.0155 \\
Morlet, unmasked & 0.6721 $\pm$ 0.0077 & 0.2954 $\pm$ 0.0080 \\
Complex Morlet, masked & 0.7013 $\pm$ 0.0262 & 0.3109 $\pm$ 0.0138 \\
Complex Morlet, unmasked & 0.6391 $\pm$ 0.0071 & 0.2413 $\pm$ 0.0094 \\
\bottomrule
\end{tabular*}
\end{table}

\subsubsection{Pixel-Level Missingness Injection}
Table~\ref{tab:ablation_pixel_mask} evaluates whether pixel-level missingness injection is useful without the explicit missingness stream.
The masked input in Figure~\ref{fig:input_representation}(b) suppresses the time-frequency regions associated with missing entries, which the unmasked input in Figure~\ref{fig:input_representation}(a) leaves intact.
For both the Morlet and Complex Morlet transforms, the masked spectrogram outperforms the unmasked spectrogram on AUROC and AUPRC.
The Morlet transform improves AUROC from 0.6721 to 0.7096 and AUPRC from 0.2954 to 0.3218, and the Complex Morlet transform shows a consistent gain from 0.6391 to 0.7013 in AUROC and from 0.2413 to 0.3109 in AUPRC.
These results indicate that the observation pattern itself carries predictive information within the spectrogram representation.

\begin{table}[t]
\centering
\caption{Effect of adding pixel-level masking and the missingness stream to the spectrogram representation.}
\label{tab:ablation_stream}
\begin{tabular*}{\columnwidth}{@{\extracolsep{\fill}}lcc}
\toprule
Setting & AUROC & AUPRC \\
\midrule
Without stream, unmasked & 0.6721 $\pm$ 0.0077 & 0.2954 $\pm$ 0.0080 \\
Without stream, masked & 0.7096 $\pm$ 0.0111 & 0.3218 $\pm$ 0.0155 \\
With stream, masked & \textbf{0.7225 $\pm$ 0.0103} & \textbf{0.3308 $\pm$ 0.0178} \\
\bottomrule
\end{tabular*}
\end{table}

\subsubsection{Explicit Missingness Stream}
Table~\ref{tab:ablation_stream} evaluates the effect of progressively adding missingness information to the spectrogram representation.
The plain spectrogram baseline does not explicitly encode observation patterns and reaches an AUROC of 0.6721 and an AUPRC of 0.2954.
Pixel-level mask injection raises these scores to 0.7096 and 0.3218 and accounts for the larger share of the overall improvement.
The explicit missingness stream takes the aligned mask image in Figure 2(c) as a separate input. 
The full CISM configuration adds the explicit missingness stream to the masked spectrogram and yields a further, complementary gain, to an AUROC of 0.7225 and an AUPRC of 0.3308.
These results support the aligned missingness stream as a representation that complements pixel-level masking rather than replaces it.

\begin{table}[t]
\centering
\caption{Effect of fusion strategies between spectrogram and missingness features.}
\label{tab:ablation_fusion}
\begin{tabular*}{\columnwidth}{@{\extracolsep{\fill}}lcc}
\toprule
Fusion & AUROC & AUPRC \\
\midrule
Concatenation & \textbf{0.7225 $\pm$ 0.0103} & \textbf{0.3308 $\pm$ 0.0178} \\
Addition & 0.6864 $\pm$ 0.0164 & 0.2996 $\pm$ 0.0208 \\
FiLM & 0.7142 $\pm$ 0.0171 & 0.3194 $\pm$ 0.0294 \\
Gating & 0.7199 $\pm$ 0.0324 & 0.3220 $\pm$ 0.0333 \\
\bottomrule
\end{tabular*}
\end{table}

\subsubsection{Spectrogram-Missingness Fusion Strategy}
Table~\ref{tab:ablation_fusion} compares strategies for fusing spectrogram and missingness features.
Concatenation achieves the best overall performance, with an AUROC of 0.7225 and an AUPRC of 0.3308.
Gating and FiLM remain competitive, whereas addition shows a clear drop on both metrics.
This pattern suggests that the two feature types should be preserved and combined through an expressive fusion mechanism, rather than merged directly through addition.
The main model therefore adopts concatenation for spectrogram-missingness fusion.

\subsubsection{Wavelet Transform Choice}

\begin{table}[t]
\centering
\caption{Effect of wavelet transform in the two-stream masked-spectrogram setting.}
\label{tab:ablation_wavelet}
\begin{tabular*}{\columnwidth}{@{\extracolsep{\fill}}lcc}
\toprule
Transform & AUROC & AUPRC \\
\midrule
Morlet & \textbf{0.7225 $\pm$ 0.0103} & \textbf{0.3308 $\pm$ 0.0178} \\
Complex Morlet & 0.7096 $\pm$ 0.0055 & 0.3304 $\pm$ 0.0093 \\
\bottomrule
\end{tabular*}
\end{table}

Table~\ref{tab:ablation_wavelet} compares Morlet and Complex Morlet transforms under the two-stream masked-spectrogram setting.
Morlet achieves higher AUROC than Complex Morlet, while AUPRC is nearly identical.
Based on this result, Morlet is used as the default transform in CISM.
Overall, the ablation results support the main design choices of the proposed framework: missingness-aware spectrogram construction, explicit missingness modeling, and learnable spectrogram-missingness fusion.

\subsection{Discussion}
The results suggest that missingness is not merely a nuisance factor in clinical time series prediction.
Pixel-level masking already accounts for much of the improvement over the plain spectrogram baseline, which indicates that observation patterns carry predictive clinical information even before any dedicated modeling.
The explicit missingness stream adds a further gain, which shows that observation patterns are most useful when they are represented as a structured signal aligned with the spectrogram, rather than only as suppressed regions in the input image.
In other words, masking and the missingness stream act on complementary levels: one constrains the input, while the other supplies an explicit, spatially aligned view of where information is absent.

The comparison with vision baselines points to a second factor behind the gain.
Generic backbones receive the same spectrogram, although they treat it as a single image in which heterogeneous variables interact before any variable-level representation is formed.
CISM instead keeps each variable band separate until patient-level aggregation, which appears well suited to clinical data in which variables differ in scale, physiological meaning, and missingness profile.
These observations should be read against the scope of the study: a single in-hospital mortality task derived from MIMIC-IV, continuous variables only, and an evaluation centered on predictive performance rather than clinical interpretability.
The evidence for variable-aligned encoding and delayed cross-variable interaction is also indirect, and rests on the contrast with vision baselines that process the spectrogram as a single image. 
A controlled ablation that varies the encoder geometry and the aggregation stage, with all other components held fixed, remains a direction for future work.

\section{Conclusion}
This work presented CISM, a Channel-Independent Spectrogram framework with a Missingness stream for clinical multivariate time series prediction. CISM addresses the variable heterogeneity and informative missingness of ICU time series: the model renders each clinical variable as a variable-wise spectrogram, preserves variable identity through variable-aligned encoding, and represents missingness through both pixel-level masking and an explicit missingness stream. On the MedMod in-hospital mortality task from MIMIC-IV, CISM attains the highest mean AUROC, AUPRC, and F1 among the time series, missingness-aware, vision, and time-frequency baselines. The ablation study attributes a large part of this gain to pixel-level masking and a further, complementary gain to the missingness stream. Beyond the mortality task studied in this work, the framework can be extended to other clinical prediction settings. A promising next step is to test whether the learned spectrogram and missingness representations align with clinically meaningful measurement decisions.

\begin{acks}
This work was supported by the Ministry of Science and ICT (MSIT), Republic of Korea, under the Global Research Support Program in the Digital Field (RS-2024-00431394), supervised by the Institute for Information \& Communications Technology Planning \& Evaluation (IITP), and by MSIT under the Advanced GPU Utilization Support Program (05-26-04-0020).
\end{acks}

\bibliographystyle{ACM-Reference-Format}
\bibliography{sypark-KDD-WS-2026}

\end{document}